\documentclass[letterpaper,journal]{IEEEtran}

\usepackage{amsmath,amsfonts}
\usepackage{amssymb}
\usepackage{algorithm}
\usepackage{algpseudocode}
\usepackage{array}
\usepackage[caption=false,font=normalsize,labelfont=sf,textfont=sf]{subfig}
\usepackage{textcomp}
\usepackage{stfloats}
\usepackage{url}
\usepackage{verbatim}
\usepackage{graphicx}
\usepackage{cite}
\usepackage{multirow}
\usepackage{booktabs}
\usepackage{newfloat}
\usepackage{listings}
\usepackage{placeins}
\usepackage{xcolor}
\usepackage[hidelinks]{hyperref}
\hypersetup{
  pdftitle={Astrolabe: Spherical-Map Guidance Across Diffusion Pipelines for Full-Body Capture from Unconstrained Images},
  pdfauthor={Shuliang Zhu, Qi Wang, Ryugo Morita, and Jinjia Zhou}
}
\newcommand{\pos}[1]{\textcolor{green!50!black}{#1}}
\newcommand{\badchange}[1]{\textcolor{red!65!black}{#1}}
\newcommand{\gain}[1]{\,{\scriptsize\pos{#1}}}
\newcommand{\loss}[1]{\,{\scriptsize\badchange{#1}}}

\newcommand{\abyes}{\textcolor{green!50!black}{\checkmark}}
\newcommand{\abno}{\textcolor{red!65!black}{$\times$}}

\floatstyle{ruled}
\newfloat{listing}{tb}{lst}{}
\floatname{listing}{Listing}


\begin{document}

\title{Astrolabe: Spherical-Map Guidance Across Diffusion Pipelines for Full-Body Capture from Unconstrained Images}

\author{Shuliang Zhu, Qi Wang, Ryugo Morita, and Jinjia Zhou%
\thanks{Preprint. This manuscript has been submitted to IEEE Transactions on Visualization and Computer Graphics for possible publication.}}

\markboth{Preprint}%
{Zhu \MakeLowercase{\textit{et al.}}: Astrolabe}

\maketitle

\begin{abstract}
Full-body capture from unconstrained photographs requires global correspondence across arbitrary views, poses, crops, and occlusions. Yet pose, geometry, and foundation features estimated in this setting are too unreliable for dense matching or appearance transfer, while diffusion rectifiers and optimization pipelines expose no common interface for consuming such uncertain correspondence. Our insight is that correspondence need not be locally accurate: its coarse viewpoint and body layout can still organize how a diffusion prior adapts and guides reconstruction. We introduce \emph{Astrolabe}, a host-portable adapter built on frozen viewpoint-guided spherical maps (SPH). A fixed bounded transform converts SPH into a spatial noise shift, which is matched during prior adaptation and reused during downstream denoising or score-distillation guidance in both pipeline categories. When a rectifier exposes a reference router, the same target/reference SPH additionally supplies coarse compatibility scores to select native appearance features; router-free optimization uses only the shared shift path. Astrolabe therefore follows one SPH--shift--adapt--guide process without dense warping or a learned control branch. Across Puzzle-IOI and 4D-Dress, it improves all reported image metrics in both hosts and all paired Puzzle-IOI geometry metrics; image gains extend to rear views, while 4D-Dress geometry remains stable overall.
\end{abstract}


\section{Introduction}
\label{sec:intro}
\IEEEPARstart{F}{ull-body} capture from unconstrained photographs must integrate identity, clothing, and body evidence scattered across unrelated images. Casual albums contain arbitrary viewpoints, poses, crops, illumination, and occlusions, so complementary observations of the same garment or body region differ in visibility and projection. A coherent avatar therefore requires \emph{global body correspondence}: relating evidence across pose and viewpoint changes so that observations distributed throughout the album support consistent target views and reconstruction.

\begin{figure*}[!tbp]
    \centering
    \includegraphics[width=1\textwidth, clip, trim=1cm 8.7cm 1.0cm 5.7cm]{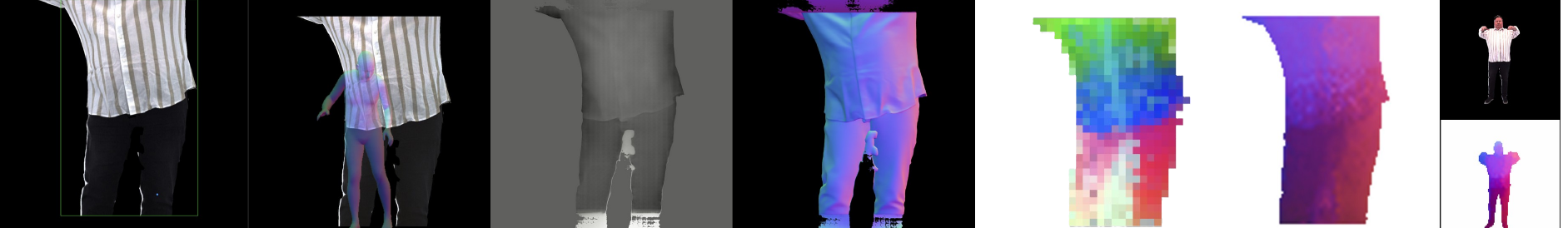}
    \caption{Candidate signals for cross-view body correspondence in unconstrained images. From left: Sapiens-derived pose, SMPL, depth, and normal estimates~\cite{khirodkar2024sapiens}; DINOv2 features; SPH; and a full-body reference. The Sapiens outputs become incomplete or locally inconsistent under truncation, loose clothing, and extreme viewpoints, while DINOv2 is semantically strong but not explicitly viewpoint-aware. SPH does not provide precise pixel correspondence, but preserves coarse viewpoint and part organization---the weak global cue that Astrolabe translates into host-native diffusion guidance.}
    \label{fig:guidance_signals}
\end{figure*}

Yet reliable global correspondence is precisely what unconstrained imagery denies. Structured reconstruction can exploit calibrated views, silhouettes, depth, or accurate body fits, while many learning-based avatars~\cite{peng2021animatable,liu2021neural,neuman2023,hu2024gaussianavatar,vid2avatar2024,exavatar2024,li2024animatable,chen2025taoavatar} rely on pose-aligned observations. Under truncation, occlusion, loose clothing, and large viewpoint changes, Sapiens-derived pose/SMPL, depth, and normal estimates~\cite{khirodkar2024sapiens} can become incomplete or locally inconsistent, while DINOv2~\cite{oquab2023dinov2} is not explicitly organized by viewpoint (Fig.~\ref{fig:guidance_signals}). Direct dense matching or appearance warping is therefore unsafe, particularly across symmetric limbs and front/back views. The challenge is to retain global, viewpoint-aware organization when local correspondence is unreliable.

Diffusion priors can complete missing views and regularize reconstruction, but do not determine how such correspondence should enter capture. \emph{Multi-view rectifiers} organize references and jointly denoise target views~\cite{up2you}, whereas \emph{optimization pipelines} such as TeCH~\cite{huang2024tech}, HaveFUN~\cite{Yang2024HaveFUN}, and PuzzleAvatar~\cite{xiu2024puzzleavatar} query a personalized prior to optimize an explicit 3D avatar. Their native interfaces differ: the former exposes joint denoising and a reference router, while the latter exposes per-render conditions and score residuals. Learned controllers such as ControlNet~\cite{Zhang2023ControlNet} and T2I-Adapter~\cite{Mou2023T2IAdapter} add architecture-specific branches and still assume a reliable condition. What is missing is a portable mechanism that carries uncertain but globally organized correspondence through variables both hosts already consume.

Our key observation is that global correspondence can guide reconstruction even when it is too weak for pixel-accurate matching. Viewpoint-guided spherical maps (SPH)~\cite{Mariotti2024SPH} preserve coarse body orientation and part layout, but are neither calibrated normals nor reliable dense correspondence. We therefore treat SPH as a weak global field rather than an appearance warp: cross-view compatibility identifies structurally plausible evidence, while spatial organization locates the diffusion bias. This retains viewpoint and symmetry information without propagating local SPH errors into reconstruction.

We introduce \emph{Astrolabe}, a complete host-portable adapter that translates weak global correspondence into native diffusion variables (Fig.~\ref{fig:pipeline}). A fixed projection and bounded transform convert SPH into a per-pixel noise shift, extending the global trajectory steering of TKG-DM~\cite{Morita2025TKGDM}; this is the common guidance path in both pipeline categories. In the router-bearing rectifier, reference and target SPH also yield coarse compatibility scores for selecting native appearance features. The two operations preserve complementary aspects of one cue: the shift carries global spatial organization, while association carries cross-view compatibility. SPH is never copied as appearance or used as a dense warp, and neither operation adds a learned control branch.

Both bindings therefore follow the same process: infer frozen SPH, translate it through fixed host-native operations, and use the guidance during capture. The rectifier applies the shift during attention-LoRA adaptation and high-noise joint denoising, while its existing router consumes SPH association. Optimization uses the same SPH-to-shift construction across reference personalization, online rendered-normal guidance, and NFDS readout; without a reference router, it has no association operation. This host-interface difference does not split the method, and both bindings retain their original appearance representation, diffusion backbone, and reconstruction procedure.

Experiments on two hosts and two datasets support this correspondence-guided design. On Puzzle-IOI, both bindings improve every paired image and geometry metric over their respective hosts. On 4D-Dress, both improve all reported image metrics while geometry remains numerically stable overall; optimization also improves P2S and surface-normal error, with only a small numerical Chamfer change. Gains extend to difficult rear views. Dense-transfer tests, association compression, and factorized controls show that accurate pixel correspondence and large association magnitudes are unnecessary; they separate association-only SPH from the coupled rectifier shift path and test spatial layout, shift matching, and guided NFDS readout in optimization. Rectifier inference remains essentially cost-neutral, while optimization is more accurate and lighter than the tested SPH- and DINO-conditioned ControlNet-color alternatives.

Our contributions are:
\begin{itemize}
    \item \textbf{(C1) Weak global correspondence for unconstrained diffusion capture.} Astrolabe is one complete adapter that translates frozen SPH into host-native guidance: a bounded spatial noise shift shared across both pipeline categories and coarse reference association in the router-bearing rectifier. It brings globally organized correspondence into multi-view rectification and optimization-based 3D capture without dense warping or a learned control branch.
    \item \textbf{(C2) A controlled account of why weak correspondence is sufficient.} Dense-transfer diagnostics, association compression, association/shift factorization, and spatial-to-mean ablations show that coarse cross-view compatibility and spatial organization remain useful without accurate pixel correspondence or large association magnitudes (Secs.~\ref{sec:exp_injection} and~\ref{sec:exp_selection}).
    \item \textbf{(C3) Cross-host evidence with lightweight deployment.} Evaluations on Puzzle-IOI and 4D-Dress demonstrate consistent image improvements across both bindings, paired Puzzle-IOI geometry gains, and stable 4D-Dress geometry; the rectifier remains essentially cost-neutral, and the optimization binding is lighter than the tested ControlNet alternative (Secs.~\ref{sec:exp_main} and~\ref{sec:exp_efficiency}).
\end{itemize}

\section{Related Work}
\subsection{Structured avatar reconstruction and correspondence}
Structured avatar capture obtains cross-view coherence from both its representation and its organized observations. Neural radiance fields and 3D Gaussian splatting provide foundations for photorealistic view synthesis~\cite{mildenhall2021nerf,kerbl20233d}; human-specific systems make them animatable through pose-conditioned radiance fields, explicit body models, or deformable Gaussian primitives~\cite{peng2021animatable,liu2021neural,shen2023x,dong2022totalselfscan,zheng2023avatarrex,moreau2024human,hu2024gaussianavatar,zielonka2023drivable,pang2024ash,li2024animatable,qian2024gaussianavatars,chen2025taoavatar}. Complementary methods separate body and clothing or model garment dynamics and physical properties~\cite{feng2022capturing,qiu2023rec,luo2024dlca,wang2023clothed,lin2024layga,su2023caphy,peng2024pica,rong2024gaussian}. Their strong fidelity relies on temporal continuity, calibrated views, or pose-aligned observations that implicitly organize evidence across the body.

Unconstrained albums remove this organization. Identity and garment evidence is distributed across unrelated poses, crops, lighting, and occlusion, while visibility and pose-to-surface alignment can disagree. A stronger avatar representation alone cannot determine how an observation of one body region should support another pose or target view. Astrolabe therefore retains the representation and reconstruction machinery of its diffusion hosts and addresses the missing evidence organization: it translates weak, viewpoint-aware global correspondence into decisions those hosts already make. This focus distinguishes the method from new rendering, deformation, or cloth representations.

\subsection{Diffusion priors for sparse and unconstrained capture}
Optimization-based capture uses diffusion priors to regularize views that sparse observations do not constrain. Score distillation converts a pretrained 2D prior into a 3D optimization signal~\cite{Poole2022DreamFusion,Wang2023ProlificDreamer}; human-specific systems add text, shape, pose, identity, or album personalization. TeCH and AvatarBooth optimize clothed avatars from text/image guidance, HaveFUN introduces few-shot pose-aware reconstruction, PuzzleAvatar assembles composable album evidence, and PFAvatar targets pose-fused outfit photographs~\cite{huang2024tech,zeng2023avatarbooth,Yang2024HaveFUN,xiu2024puzzleavatar,Xi2025PFAvatar}. These methods provide strong generative regularization, but the optimization signal still needs a reliable way to relate subject evidence to the current rendered viewpoint.

Rectification and multi-view generation organize observations before or alongside reconstruction. Morphable Diffusion, Human-3Diffusion, DreamAvatar, and MVDream learn multi-view or shape-aware priors~\cite{Chen2024MorphableDiffusion,Xue2024Human3Diffusion,Cao2024DreamAvatar,mvdreambooth2024}; DiffusionAvatars, Cap4D, and GAF couple diffusion with head, portrait, or video-avatar representations~\cite{Kirschstein2024DiffusionAvatars,taubner2025cap4d,tang2025gaf}. UP2You instead rectifies an unconstrained collection through joint target denoising and pose-correlated routing of native reference features~\cite{up2you}. This host must select compatible evidence immediately, whereas optimization hosts expose per-render score residuals rather than a reference router. Astrolabe translates one weak correspondence cue into these native interfaces: the spatial shift is shared across both host categories, and the same cue additionally supplies coarse association only when an existing router can consume it.

\subsection{Conditional and reference-guided diffusion}
General diffusion controllers inject conditions through learned model-side interfaces. ControlNet and T2I-Adapter attach spatial-conditioning branches, IP-Adapter introduces decoupled image-prompt attention, and GLIGEN grounds generation through gated spatial features~\cite{Zhang2023ControlNet,Mou2023T2IAdapter,ye2023ipadapter,Li2023GLIGEN}. These mechanisms are effective when a stable condition and a particular denoiser can be trained together, but their interfaces depend on network activations or attention modules and presume that the supplied condition is sufficiently aligned.

Human synthesis specializes these interfaces for skeletons, poses, and cross-view appearance. PoCoLD, HumanSD, CFLD, Stable-Pose, and GRPose encode structured pose constraints~\cite{Han2023PoCoLD,Ju2023HumanSD,Lu2024CFLD,wang2024stable,yin2025grpose}; cross-view masked diffusion and DreamPose transfer appearance across viewpoints or frames~\cite{pham2024cross,Zhang2023DreamPose}, while domain-adapted human controls improve shape and pose specification~\cite{buchheim2025controlling}. Such systems generally learn an aligned condition jointly with one model family. Astrolabe instead treats correspondence as frozen and uncertain: it maps SPH to variables the host already consumes---diffusion noise and, where available, reference-selection scores. The rectifier's small LoRA only adapts existing attention to the shifted-noise family; it neither encodes SPH nor forms a separate conditioning interface. The cue therefore changes trajectory layout and which native appearance features are read without becoming an appearance encoder or requiring a shared learned control branch.

\subsection{Weak correspondence and noise-domain steering}
Strong geometric coordinates are valuable only when their estimates remain reliable. Canonical body models and dense human coordinates provide explicit semantics but inherit fitting and correspondence errors in unconstrained imagery~\cite{loper2023smpl,Guler2018DensePose}; foundation features improve semantic matching and human perception without guaranteeing viewpoint-aware body correspondence~\cite{oquab2023dinov2,khirodkar2024sapiens}. SphericalMaps instead augments self-supervised features with weak viewpoint-aware geometry for symmetric correspondence~\cite{Mariotti2024SPH}. Astrolabe deliberately consumes this signal below the level of dense transfer: coarse compatibility ranks native reference evidence, and spatial organization lays out trajectory guidance. This preserves the useful global relation without treating local SPH matches as trustworthy appearance correspondence.

Noise steering provides a branch-free destination for that spatial organization. TKG-DM shows that global channel-mean shifts can steer generated color without modifying the denoiser~\cite{Morita2025TKGDM}. Astrolabe connects this noise-domain premise to weak global correspondence by extending the fixed transformation to a bounded per-pixel field; in the router-bearing binding, the same frozen cue additionally ranks native appearance values without replacing them. The resulting contribution is neither a new dense matcher nor another learned controller, but one cue-to-host conversion whose two boundaries---coarse rather than dense selection, and spatial rather than channel-mean guidance---are tested directly in our ablations.

\section{Preliminaries}
\subsection{Noise shifting for diffusion control}
\label{sec:prelim_shift}
With $\epsilon_T\!\sim\!\mathcal{N}(0,I)$ and latent channels $c\!\in\!\{1,\ldots,C\}$, TKG-DM~\cite{Morita2025TKGDM} adds a channel-wise bias
\begin{equation}
\epsilon'_{c,h,w}=\epsilon_{c,h,w}+s_c,
\label{eq:mean_shift}
\end{equation}
which steers the denoising trajectory while leaving per-channel variance unchanged. Two properties make this interface useful here: the shift acts from a high-noise state, and its energy can remain small relative to scheduler noise. TKG-DM instantiates $s$ from a target color through a fixed color-to-channel map. Astrolabe instead derives a spatial field $s_{c,h,w}$ from SPH, allowing the cue's global layout to enter variables that diffusion hosts already consume. This direct, branch-free interface remains available whether or not a host exposes explicit cross-view routing.

\subsection{Spherical maps as weak global correspondence}
Viewpoint-guided spherical maps (SPH)~\cite{Mariotti2024SPH} map each visible token to a learned unit vector $q\in\mathbb S^2$. The field preserves coarse viewpoint and part layout, but is neither a calibrated camera/world normal map nor reliable pixel correspondence. Both pipeline categories use one frozen DINOv2-B/14-backed checkpoint: the rectifier evaluates it at $896^2$ to obtain a $64^2$ token grid, while optimization uses the same predictor through offline RGB-reference and online rendered-normal paths. Host-provided foreground masks suppress invalid tokens. Rectifier runs use binary support without extra shift-mask erosion or softening; optimization personalization intersects offline body support with the current training-asset support, while distillation uses renderer masks. SPH is therefore deliberately used at coarse granularity: it is structured enough to lay out a shift and, where available, rank compatible regions, but too uncertain to justify dense appearance transfer.

\subsection{Two diffusion hosts}
\label{sec:prelim_hosts}
The two hosts expose different decisions through which global correspondence can guide capture. The \textbf{rectifier host} adapts attention layers, routes reference features, and jointly denoises target views~\cite{up2you}; the \textbf{optimization host} personalizes a 2D prior and queries it one render at a time to optimize DMTet through score-distillation updates~\cite{xiu2024puzzleavatar}. Astrolabe translates the same frozen SPH cue into the variables each host already consumes: a spatial noise field in both, and coarse compatibility scores in the rectifier's existing router. The shared object is therefore the cue translation, not identical trained weights or host operations.

\paragraph{Timestep convention.}
We use the forward DDPM index $t\in\{0,\ldots,T-1\}$, where larger $t$ means more noise; a normalized interval $[a,b]$ denotes $t\in[aT,bT)$. ``Early'' reverse denoising therefore means large $t$. Optimization progress is $p=i/N$.

\section{Method}
\label{sec:method}

Astrolabe translates frozen weak global correspondence into the native guidance variables of diffusion-based full-body capture. Given reference images and a target or current observation, the frozen SphericalMaps predictor~\cite{Mariotti2024SPH} produces SPH fields whose coarse viewpoint organization is useful even though their local matches are unreliable. Astrolabe preserves two complementary properties of this cue: global spatial organization, realized as a bounded latent-noise field, and cross-view compatibility, realized as coarse association when a host exposes a reference router.

\begin{figure*}[!tbp]
    \centering
    \includegraphics[width=\textwidth, clip, trim=0cm 4.1cm 0cm 5.1cm]{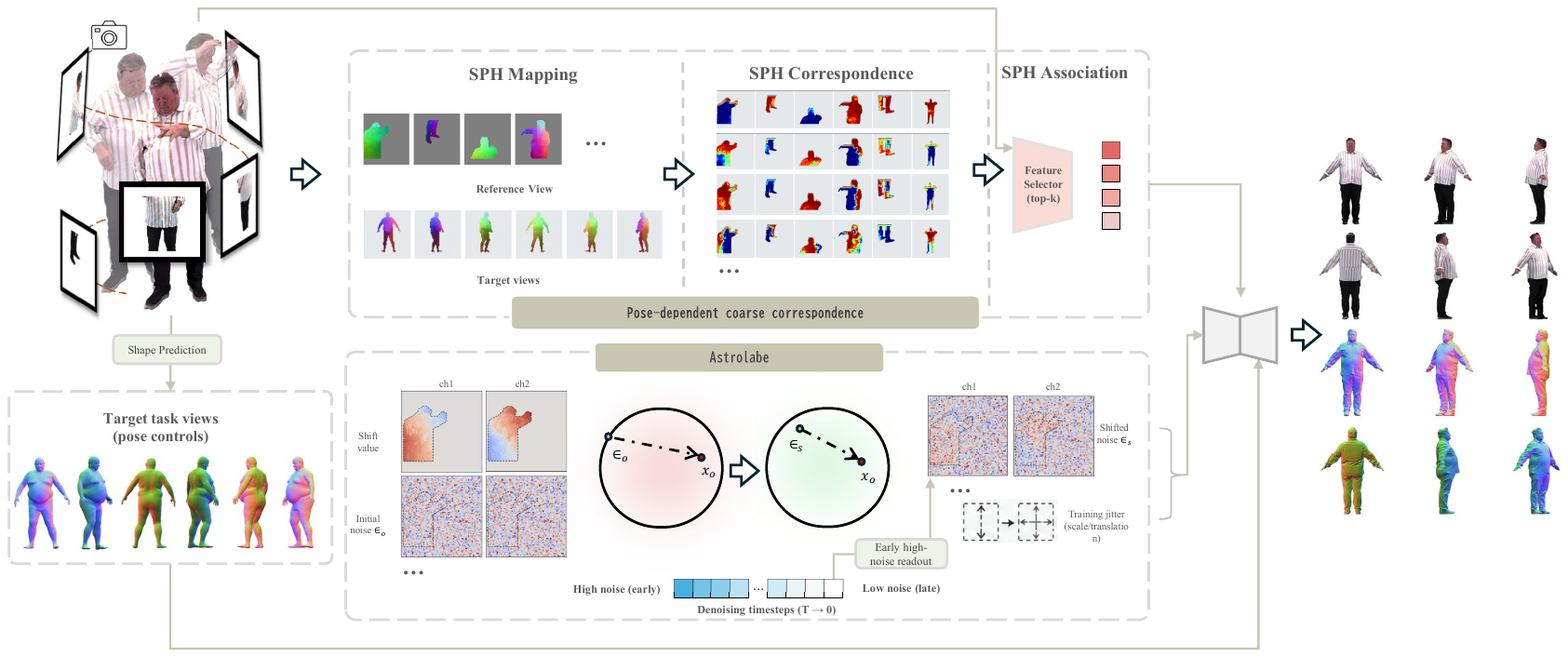}
    \caption{\textbf{From weak global correspondence to host-native guidance.} The frozen SphericalMaps predictor extracts SPH from references and target/current observations. Astrolabe preserves its global spatial organization as a fixed bounded field $\delta$, whose scaled realization moves $\epsilon_o\!\to\!\epsilon_s$ in both host categories. In the rectifier, shift-matched LoRA adaptation and inference-time initialization/readout form one coupled shift path; reference/target SPH additionally provides coarse compatibility for top-$k$ selection of native appearance features. Reverse denoising proceeds from $t=T-1$ toward $0$; scale/translation jitter is used only during optimization personalization.}
    \label{fig:pipeline}
\end{figure*}

Figure~\ref{fig:pipeline} shows this cue-to-host translation. Every binding predicts SPH, converts its spatial layout through a fixed projection and bounded nonlinearity into $\delta$, and realizes a host-scaled shift $s$. In the rectifier, $s$ defines one coupled path across shift-matched attention-LoRA adaptation, initialization, and high-noise readout; the LoRA is the host's trainable carrier of this path, not another adapter output. Reference/target SPH additionally supplies coarse association to the existing router. The router-free optimization binding carries the same spatial construction through personalization and 3D guidance. These host bindings are coordinated realizations of one correspondence cue rather than separate methods or contributions.

Three constraints keep the translation faithful to a weak cue. First, SPH controls only layout and compatibility; selected appearance values remain native host features. Second, the fixed bounded map steers rather than overwrites the pretrained trajectory. Third, Astrolabe reuses routing scores, noise tensors, and residuals already exposed by the hosts instead of adding a learned control branch. Section~\ref{sec:shift} constructs the spatial field, Sec.~\ref{sec:assoc} constructs coarse association, and Sec.~\ref{sec:hosts} binds these outputs to the two capture pipelines while retaining their original appearance representations and reconstruction machinery~\cite{up2you,xiu2024puzzleavatar}.

\subsection{Spatial realization of weak correspondence}
\label{sec:shift}
Astrolabe carries SPH's global spatial organization into diffusion through a bounded latent-noise field. Building on noise-domain steering~\cite{Morita2025TKGDM}, the construction converts the coarse correspondence cue into a per-pixel bias rather than a dense warp, a learned control branch, or one global channel shift. It is the shared adapter output in both pipeline categories and remains usable when no reference router exists.

The frozen predictor emits $q=(q_1,q_2,q_3)$ in a learned basis with no assigned physical axes. We preserve this checkpoint-defined order, encode $I_{\mathrm{sph}}=(q+1)/2=(R,G,B)$, and center it as $\bar I_{\mathrm{sph}}=I_{\mathrm{sph}}-0.5$. Let
\begin{equation}
P=\begin{bmatrix}
0&0&0&1\\0&0&1&0\\0&1&0&0\\1&0&0&0
\end{bmatrix}
\label{eq:channel_reverse}
\end{equation}
reverse the four mapped latent channels. The fixed projection is
\begin{equation}
M=\begin{bmatrix}0&0&0\\-2&0&2\\1&1&-2\\0&0&0\end{bmatrix},\qquad
u=\tfrac{1}{4}PM\bar I_{\mathrm{sph}}.
\label{eq:fixed_map}
\end{equation}
Following TKG-DM's empirical color-to-latent design~\cite{Morita2025TKGDM}, we algebraically adapt the projection to the four-channel latent order without fitting evaluation subjects. The zero-sum nonzero rows of $M$ cancel common SPH-channel offsets; $P$ only permutes channels, and $1/4$ limits pre-nonlinearity amplitude. Because the SPH basis has no calibrated axes, individual rows carry no body-part or world-coordinate semantics.
After bicubic resampling to the latent grid, the canonical $[0,1]$-encoded construction is
\begin{equation}
\delta=0.8\,\tanh\!\left(2\,\operatorname{atanh}
\big(\operatorname{clip}(u,-0.999,0.999)\big)\right)\odot m,
\label{eq:sph2shift}
\end{equation}
\begin{equation}
s=\kappa\delta,\qquad \epsilon_s=\epsilon_o+s,\qquad
\epsilon_o\sim\mathcal{N}(0,I),
\label{eq:scaled_shift}
\end{equation}
where $m$ is host-provided foreground support and $\kappa$ is host/regime-specific. The bounded $\delta$ is the shared representation and $s$ the realized host input. Both hosts share $M$, $P$, the foreground-masking operation, and the nonlinearity but retain native masks and input conventions: Eq.~\eqref{eq:sph2shift} applies directly to the rectifier and online distillation, whereas personalization's $[-1,1]$ loader induces a fixed $2{\times}$ pre-nonlinearity scale absorbed into its reported $\kappa$, since $M$'s nonzero rows sum to zero.

The fixed construction prevents a weak cue from becoming an architecture-specific appearance encoder. Matrices $M$ and $P$ and the nonlinear transform are unchanged throughout all experiments, so neither host learns an SPH-to-latent projection or assigns physical axes to the checkpoint basis. After host-specific resampling, $\delta\in\mathbb R^{4\times H_z\times W_z}$ matches a latent-noise sample, and foreground masking leaves background locations unchanged. Unlike feature conditioning, $\delta$ is never concatenated with a UNet activation; the realized $s$ enters a noise tensor or residual already consumed by the host.

\begin{figure}[!tbp]
    \centering
    \includegraphics[width=0.96\linewidth, clip, trim=4.0cm 5.0cm 4.5cm 5.4cm]{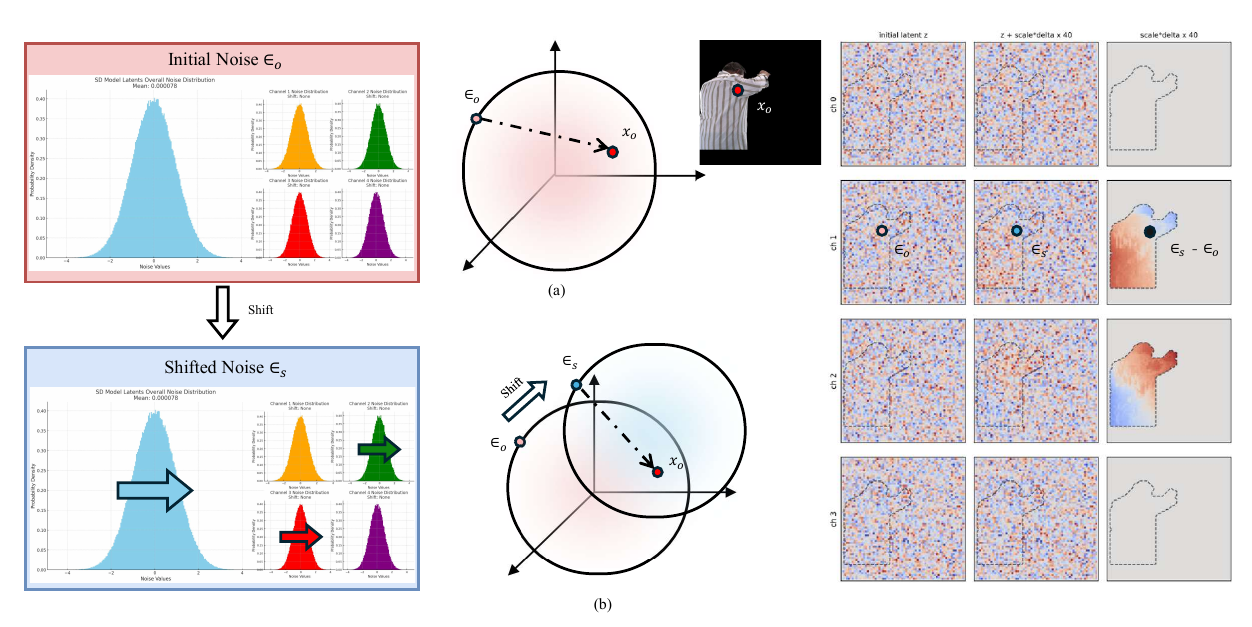}
    \caption{\textbf{Spatial realization of weak correspondence.} Left: overall and per-channel histograms of $\epsilon_o$ and $\epsilon_s$; small shifts move the means while remaining empirically close to Gaussian, without implying an exact global Gaussian. Middle: $\epsilon_o\!\to\!\epsilon_s$ relocates the trajectory's starting state toward $x_o$. Right: per-channel $\epsilon_o$, $\epsilon_s$, and $s=\epsilon_s-\epsilon_o$ (shown at $\times40$) reveal an SPH-aligned spatial field rather than a global bias.}
    \label{fig:shift_mechanism}
\end{figure}

The bounded transform amplifies weak nonzero entries without allowing large ones to grow unbounded. For $|u|<1$, $\tanh(2\operatorname{atanh}u)=2u/(1+u^2)$. At each location, $\epsilon_s$ is therefore a unit-variance Gaussian with mean $s(c,h,w)$, but aggregating a spatially varying field yields a mixture rather than one globally translated Gaussian. Because the realized shifts are small, the empirical channel histograms in Fig.~\ref{fig:shift_mechanism} remain close to Gaussian. We treat this as an empirical approximation, not distributional invariance.

\begin{figure}[!tbp]
    \centering
    \includegraphics[width=0.96\linewidth, clip, trim=8cm 5.5cm 8cm 2.5cm]{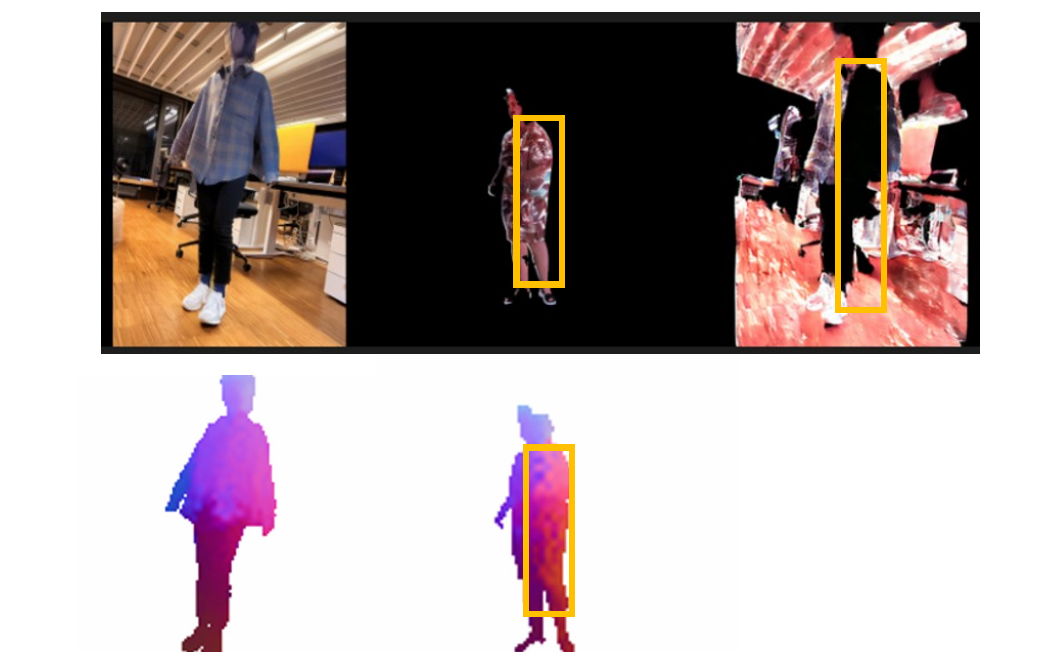}
    \caption{\textbf{Local response to the correspondence-derived shift.} Top: an in-the-wild photograph, its masked subject, and the transfer result; bottom: source and target SPH maps. The highlighted target leg has no source counterpart and remains intact, indicating that the host responds to the layout of $s$ rather than globally copying appearance.}
    \label{fig:local_transfer}
\end{figure}

\paragraph{Why spatial (Fig.~\ref{fig:shift_mechanism}, Fig.~\ref{fig:local_transfer}).}
Spatial variation preserves \emph{where} the correspondence cue should act, which a global channel mean cannot express. The per-pixel field retains coarse viewpoint and part layout without asserting local matches. In Fig.~\ref{fig:local_transfer}, a target leg segment with no source counterpart remains intact, indicating a localized response rather than global appearance copying. The spatial-to-mean optimization control directly tests this layout, while the factorized rectifier control compares the complete coupled shift path against association-only SPH when explicit routing is available (Sec.~\ref{sec:exp_injection}).

\subsection{Coarse correspondence for native reference selection}
\label{sec:assoc}
The rectifier must decide which scattered reference evidence is compatible with each target view, even though SPH is too uncertain for dense transfer. Astrolabe therefore uses SPH only to rank the host's native reference features. The cue determines \emph{which features to read}, while the selected values still come from the host's learned appearance representation; no SPH value is copied as appearance or used to warp a reference image.

Given valid frozen-SPH outputs $q^r_j,q^t_\ell$ for reference token $j$ and target token $\ell$, Astrolabe first computes target-specific compatibility. We encode $c=(q/\|q\|_2+1)/2$ and use $d_{j\ell}=\|c^r_j-c^t_\ell\|_2$; DINOv2-B remains internal to the SPH predictor, and its appearance features do not enter this distance. Let $d_k(j)$ be the mean of the $k=4$ smallest target-token distances and $d_{.60}(j)=Q_{.60}(\{d_{j\ell}\}_{\ell})$. The reference-token score is
\begin{equation}
a_j=\lambda\big(d_{.60}(j)-d_k(j)\big)-d_k(j),\qquad \lambda=0.45.
\label{eq:assoc_score}
\end{equation}
The $-d_k(j)$ term rewards a compatible target region, while $d_{.60}(j)-d_k(j)$ favors a match distinctive from that token's typical distance. Equation~\eqref{eq:assoc_score} produces one score per reference token for the current target view; it neither selects a target pixel nor defines a warp. Joint scoring over valid tokens from all references also prevents each reference from independently rescaling itself into an equally strong candidate.

Because raw compatibility varies across target views, we normalize it robustly over valid reference tokens. With linear quantile interpolation, let $\ell=Q_{.35}(a)$ and $h=Q_{.98}(a)$, and compute $r=\operatorname{clip}((a-\ell)/\max(h-\ell,10^{-6}),0,1)^{1.2}$. We set $\widehat r=r$ for $r\leq0.55$ and $\widehat r=0.55+0.30(r-0.55)$ otherwise, followed by $w=4.85\widehat r$; invalid tokens remain zero. The lower quantile suppresses diffuse matches, the upper quantile prevents outliers from fixing the range, and the reduced high-value slope limits extreme scores.

Thin, identity-sensitive head boundaries can disappear during foreground downsampling. Within each reference bounding box, we therefore define the head as pixels with normalized vertical coordinate below $0.28$, dilate its inner mask boundary by one feature pixel, and floor that boundary at $0.65$ of the map's positive maximum. This support operation protects hair and head evidence after masking without introducing facial correspondence; SPH still determines the remaining ranking.

The final selector limits score extremes before allocating its fixed token budget. For nonempty positive support, let $W=\max_{w>0}w$, with knees $l=0.22W$ and $h_s=0.66W$ and widths $\sigma_l=0.11W$ and $\sigma_h=0.15W$; we apply the monotone soft-clip
\begin{equation}
\operatorname{SC}(w)=
\begin{cases}
[l-\sigma_l(1-e^{-(l-w)/\sigma_l})]_+, & 0<w<l,\\
w, & l\leq w\leq h_s,\\
h_s+\sigma_h(1-e^{-(w-h_s)/\sigma_h}), & w>h_s,\\
0, & w=0.
\end{cases}
\label{eq:softclip}
\end{equation}
Here $W$ is computed on the current association tensor. Structural zeros remain zero, while the exponential tails preserve positive ordering and limit leverage near either extreme. From the $N_rH_fW_f$-token pool, the selector retains the $H_fW_f$ highest-weight tokens---one feature-map equivalent, with no random tokens---and passes their native values to the router. SPH therefore supplies eligibility and priority, whereas the payload remains the host's appearance feature; an all-zero map contributes zero features.

\paragraph{Selection, not loudness.}
A weak correspondence cue should control coarse eligibility and ranking without relying on large magnitudes. To compress contrast while preserving positive support, let $\mathcal S=\{j:w_j>0\}$ and $\bar w_{\mathcal S}=|\mathcal S|^{-1}\sum_{j\in\mathcal S}w_j$. Before top-$k$, we apply
\begin{equation}
w'_j =
\begin{cases}
\beta w_j+(1-\beta)\bar w_{\mathcal S}, & j\in\mathcal S,\\
0, & j\notin\mathcal S,
\end{cases}.
\label{eq:value_blend}
\end{equation}
The forward order is fixed: soft-clip $w$, blend on the native $64^2$ grid, bilinearly resize to each reference-feature resolution, apply the resized reference mask, and select top-$k$. If $\mathcal S=\emptyset$, blending is skipped and the contributed features remain zero despite implementation-dependent tie indices. For $0<\beta\leq1$, Eq.~\eqref{eq:value_blend} is strictly monotone on $\mathcal S$, preserving native-grid support and ranking while compressing magnitudes. Resizing can alter boundary order before top-$k$, and $\beta=0$ ties all positive native values, so final membership is not claimed to be identical. The main model uses $\beta=0.2$, where the gain persists under strong compression (Sec.~\ref{sec:exp_selection}); this qualified robustness is what we call ``selection, not loudness.''

\subsection{Host-native integration across capture pipelines}
\label{sec:hosts}
Astrolabe binds one correspondence adapter according to the variables exposed by each host rather than forcing identical internal operations. Both bindings use the frozen SPH predictor and fixed spatial construction. The rectifier couples shift-matched adaptation with inference-time trajectory guidance and additionally feeds coarse association to its router; optimization carries the spatial field through personalization and score-distillation updates without association. Host-specific scales and timestep gates respect each native schedule, and neither binding changes the host's avatar representation or reconstruction procedure.

\paragraph{Rectifier binding}
The rectifier binding follows \texttt{up2you}'s shape/pose processing, latent encoder, pose-correlated aggregator, joint denoiser, native reference values, and schedule~\cite{up2you}. First, its preliminary pose-conditioned target RGB and the input photographs provide masked target and reference SPH. Their coarse compatibility from Sec.~\ref{sec:assoc}, including Eq.~\eqref{eq:value_blend}, replaces only the router scores; selected payloads remain the host's learned appearance features. Second, the spatial field defines a single shift path from adaptation to inference: a small attention LoRA is trained with shifted noise while the remaining backbone stays frozen, then the same construction perturbs initial target noise at scale $0.7$ and supplies scale-$0.5$ readout for $t>600$, with later reverse steps unshifted. This LoRA exists to match the denoiser to the shifted-noise family and is not an independent conditioning module. Consequently, the association-only diagnostic disables both shift uses and the trained LoRA, whereas every active rectifier shift path includes them as one coupled mechanism. The effective shift is approximately $0.04$, so trajectory guidance remains weak while association directly organizes cross-view evidence.

\paragraph{Optimization binding (SDS)}
\label{sec:host_opt}
The optimization binding follows PuzzleAvatar's personalized 2D prior, UNet/VAE, renderer, NFDS/CFG, sampling, and two-phase DMTet optimization~\cite{xiu2024puzzleavatar,gao2022get3d}. Because this host has no reference-token pool or router, it cannot consume explicit association. Astrolabe instead carries SPH's global layout through an offline reference-RGB field during personalization and an online rendered-normal field during DMTet optimization. The representation and reconstruction schedule remain unchanged, and no conditioning branch is added.

\paragraph{Robustness to cue--observation misalignment.}
Weak SPH and observation masks can be slightly misaligned during personalization. We therefore perturb only the input-side shift with isotropic scale $\gamma\!\sim\!\mathcal U(0.9,1.1)$ and normalized-grid translations $\mathcal U(-0.03,0.03)$, using bilinear resampling and zero padding. The aligned target shift and all RGB, normal, and mask targets remain unchanged. This controlled mismatch discourages dependence on one exact alignment without geometrically altering supervision.

\paragraph{Shift-matched personalization.}
The personalized denoiser must learn the same correspondence-derived noise family that later guides 3D reconstruction. For RGB reference $x$, normal image $n$, and VAE latents $z=E(x),z_n=E(n)$, define
\begin{equation}
\widetilde z_t(z,\eta)=\sqrt{\alpha_t}z+\sqrt{1-\alpha_t}\,\eta.
\label{eq:noisy_latent}
\end{equation}
One offline SPH field supplies both slots. We jitter only its input realization $s_{\rm in}$ while retaining aligned $s_{\rm tgt}$ as the prediction target:
\begin{equation}
\begin{aligned}
\mathcal{L}_{\rm match}=\mathbb{E}\!\big[&
\|\epsilon_\theta(\widetilde z_t(z,\epsilon+s_{\rm in}),y)
-(\epsilon+s_{\rm tgt})\|_2^2\\
&+\|\epsilon_\theta(\widetilde z_t(z_n,\epsilon+s_{\rm in}),y)
-(\epsilon+s_{\rm tgt})\|_2^2\big].
\end{aligned}
\label{eq:shift_match_loss}
\end{equation}
Each real RGB reference is processed once by the frozen SPH predictor, and its masked field supplies the aligned target to both RGB and normal slots. Equation~\eqref{eq:shift_match_loss} converts standard noise prediction into shift matching without adding an output branch: the denoiser receives the jittered realization but predicts the aligned one. The shift scale is $0.8$ for $120\leq t<300$.

\paragraph{Online correspondence during distillation.}
The cue is recomputed as the avatar changes. During both DMTet phases, the frozen SPH predictor produces $I_{\rm sph}={\rm SPH}_{\rm pred}(n)\odot m$ from rendered normal $n$ treated as a three-channel image. This is a learned prediction rather than an analytic normal-to-sphere conversion. Geometry is frozen during appearance, which uses the lower shift scale and no RGB--SPH mixture.

\paragraph{Conditioning and guided NFDS readout.}
The online correspondence field influences both the sampled state and the residual read by NFDS. At camera $c$, let $x=R(\phi;c)$, $z=E(x)$, and $\widetilde z_t=\widetilde z_t(z,\epsilon+s_{\rm cond}(t))$. Denote conditional, unconditional, and negative-prompt UNet predictions by $\hat\epsilon_y,\hat\epsilon_\emptyset,\hat\epsilon_{y^-}$. The NFDS/CFG residual is
\begin{equation}
g_z=(1-\alpha_t)
\left[w_g(\hat\epsilon_y-\hat\epsilon_\emptyset)+r_t\right].
\label{eq:sds}
\end{equation}
\begin{equation}
r_t=g_{\rm grad}(t)(\hat\epsilon_\emptyset-s_{\rm grad}(t))
+[1-g_{\rm grad}(t)](\hat\epsilon_\emptyset-\hat\epsilon_{y^-}),
\label{eq:nfds}
\end{equation}
where binary $g_{\rm grad}(t)$ activates SPH readout; outside its support, $r_t$ reverts to the host's negative-prompt residual. Independently gated versions of the same $\delta$ therefore control noising and explicit gradient readout.

Separate gates let the cue change the sampled state and the NFDS residual over different timestep supports. In geometry,
\begin{equation}
s_{\rm cond}(t)=0.2\delta\,\mathbf1_{[40,300)}(t),\qquad
s_{\rm grad}(t)=0.2\delta\,\mathbf1_{[80,300)}(t).
\label{eq:geom_gates}
\end{equation}
Appearance multiplies both paths by $0.35$, uses condition support $[0,1000)$, and restricts gradient readout to $[0,200)$. Geometry therefore emphasizes the noisy portion of its restricted schedule, whereas appearance permits broad conditioning but confines explicit residual guidance to low indices. These scales and supports are host/regime bindings, not learned parts of the shared map.

The guided residual reaches DMTet through the differentiable VAE input path rather than an image-space shortcut. With detached UNet predictions and frozen VAE weights, the host injects row-wise norm-clipped $\mathcal C_{0.1}(g_z)$ on $z$:
\begin{equation}
\nabla_\phi\mathcal L_{\rm SDS}
=\left(\frac{\partial z}{\partial\phi}\right)^{\!\top}
\mathcal C_{0.1}(g_z),\qquad
\frac{\partial z}{\partial\phi}
=\frac{\partial E(x)}{\partial x}
\frac{\partial R(\phi;c)}{\partial\phi}.
\label{eq:sds_chain}
\end{equation}
The UNet prediction is detached, but $E(x)$ is not. Equation~\eqref{eq:sds_chain} therefore preserves the gradient path from the injected latent residual through the frozen VAE encoder to the renderer.

\paragraph{Spatial-layout control.}
To isolate spatial organization from per-channel shift magnitude, the ablation in Table~\ref{tab:t3_optim} averages bounded, masked $\delta$ before phase scaling and timestep gates:
\begin{equation}
\Delta_{b,c}=\frac{\sum_{h,w}m_{b,h,w}\delta_{b,c,h,w}}
{\max(\sum_{h,w}m_{b,h,w},1)},\qquad
\delta^{\rm mean}_{b,c,h,w}=\Delta_{b,c}m_{b,h,w}.
\label{eq:channel_mean}
\end{equation}
Thus $\delta^{\rm mean}$ preserves each foreground channel mean while removing only its spatial layout.

\begin{algorithm}[!tbp]
\caption{Astrolabe on the optimization host}
\label{alg:shift_sds}
\begin{algorithmic}[1]
\Require UNet $\epsilon_\theta$, renderer $R$, ${\rm SPH}(\cdot)$, fixed map $M$, schedule $\{\alpha_t\}$, gates $g_{\rm cond},g_{\rm grad}$
\For{$\text{phase}\in\{\mathrm{GEOM},\mathrm{APP}\}$}
  \State freeze appearance (GEOM) / geometry (APP)
  \For{$i=1$ to $N_{\text{phase}}$}
    \State sample camera $c$ and timestep $t$
    \State render $(x,n,m)\leftarrow R(\phi;c)$; $I_{\rm sph}\leftarrow{\rm SPH}_{\rm pred}(n)\odot m$
    \State $\delta\leftarrow$ Eqs.~\eqref{eq:fixed_map}--\eqref{eq:sph2shift}; scale and gate it into $s_{\rm cond},s_{\rm grad}$
    \State $\widetilde z_t\leftarrow\operatorname{add\_noise}(E(x),\epsilon+s_{\rm cond},t)$
    \State form $g_z$ by Eqs.~\eqref{eq:sds}--\eqref{eq:nfds}; inject $\mathcal C_{0.1}(g_z)$ at $z=E(x)$ through Eq.~\eqref{eq:sds_chain}; update $\phi$
  \EndFor
\EndFor
\end{algorithmic}
\end{algorithm}

Algorithm~\ref{alg:shift_sds} summarizes the optimization binding after personalization. Each iteration renders the current avatar, predicts SPH from its normal map, constructs $\delta$ with the fixed cue translation, applies phase-specific condition/readout gates, and injects the latent residual into the DMTet update. Geometry and appearance differ only in trainable variables, scales, and supports. Both host bindings therefore begin with the same frozen correspondence representation and fixed spatial construction; their endpoints differ because one guides joint denoising and native reference routing, whereas the other guides an SDS residual without a reference-token pool.

\section{Experiments}
\label{sec:experiments}
Our experiments test whether weak global correspondence can improve unconstrained full-body capture without reliable dense matching. We ask three questions: (i) does the complete adapter improve both rectification and optimization across datasets and difficult rear views (C1, C3); (ii) how do coarse association and spatial trajectory guidance carry cross-view compatibility and global layout into each host (C2); and (iii) do the gains persist without accurate warping, large association magnitudes, or a learned control branch (C2, C3)? We report the protocol, overall and difficult-view results, factorized mechanisms, correspondence diagnostics, and deployment cost.

\begin{figure*}[!t]
    \centering
    \includegraphics[width=0.96\textwidth,clip,trim=0.2cm 4.6cm 0.5cm 4.3cm]{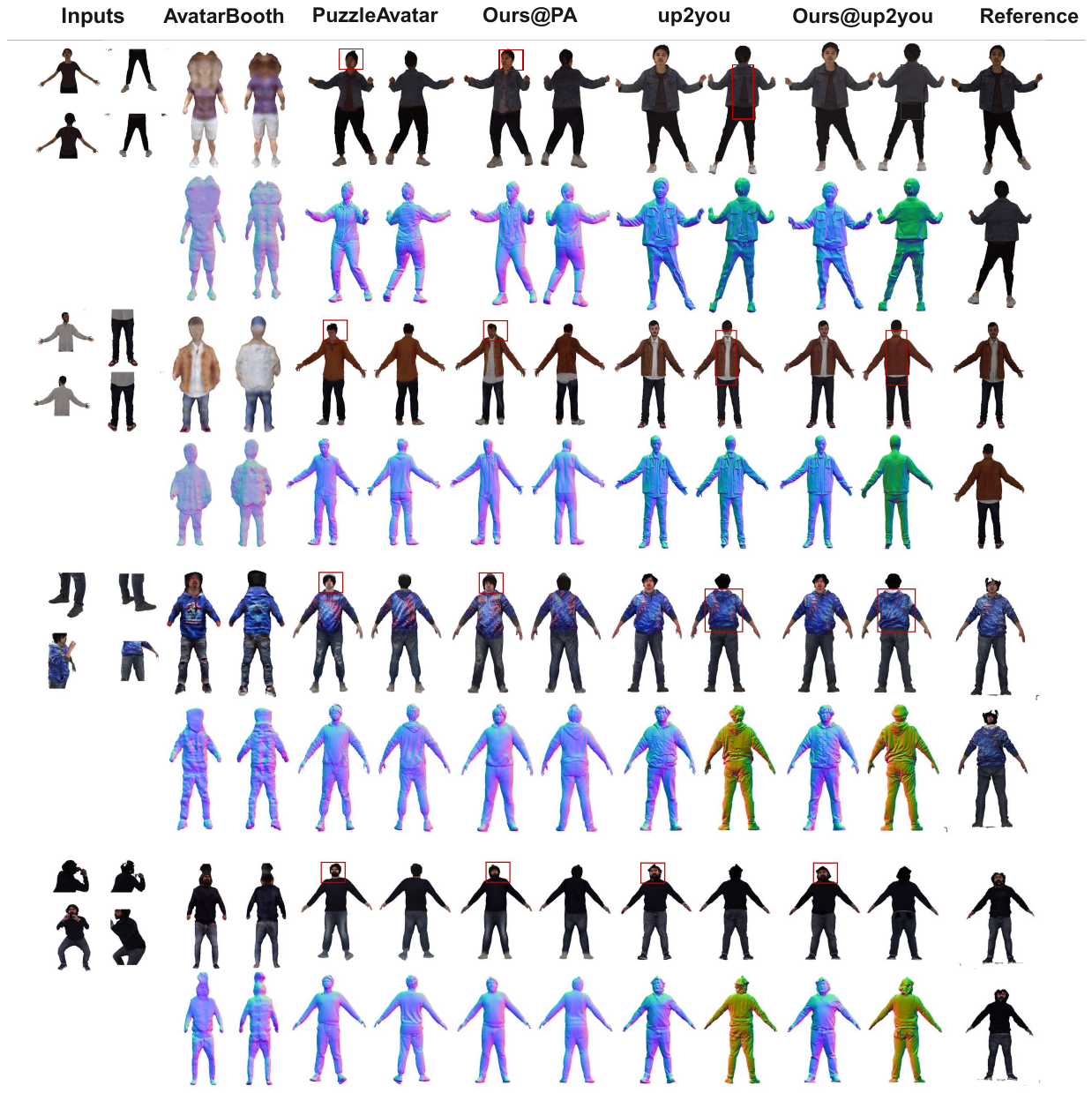}
    \caption{\textbf{Qualitative comparison across both host families.} Embedded ``Ours'' labels denote Astrolabe. Columns compare AvatarBooth, PuzzleAvatar, Astrolabe@PA, \texttt{up2you}, and Astrolabe@\texttt{up2you}; each reconstruction is shown with front/back RGB and normal views. Puzzle-IOI cases emphasize rear appearance and garment layout. In the two matched-input 4D-Dress studies, all methods use the same $16$ references (eight partial base and eight full-body update views); Table~\ref{tab:t1_main}'s rectifier block instead reports the separate ten-case aggregate, with eight partial references per case.}
    \label{fig:qual_main}
\end{figure*}

\begin{table*}[!t]
\centering
\caption{\textbf{Main quantitative comparison.} 2D metrics use independently black-composited prediction/GT RGBA, full-image PSNR/SSIM, and scalar Alex-LPIPS. Rectifier meshes are reconstructed from six generated targets; Puzzle-IOI uses $10$ subjects with $12$ references, and both 4D-Dress host blocks use $10$ cases. The paired Puzzle-IOI optimization comparison uses nine subjects; AvatarBooth covers eight and is reported as a non-paired mean. Parentheses give Astrolabe's paired or within-protocol change (green better, red worse). Normal$^{\dagger}$ is angular surface-normal error in degrees; Normal$^{\ddagger}$ is the unitless native PuzzleAvatar normal-image error and is not numerically comparable.}
\label{tab:t1_main}
\setlength{\tabcolsep}{4.5pt}
\renewcommand{\arraystretch}{1.08}
\resizebox{\textwidth}{!}{%
\begin{tabular}{l l l ccc ccc}
\toprule
 & & & \multicolumn{3}{c}{\textbf{3D (mesh)}} & \multicolumn{3}{c}{\textbf{2D (multi-view/rendered)}} \\
\cmidrule(lr){4-6}\cmidrule(lr){7-9}
\textbf{Host} & \textbf{Dataset} & \textbf{Method} & Chamfer$\downarrow$ & P2S$\downarrow$ & Normal$^{\dagger/\ddagger}\downarrow$ & PSNR$\uparrow$ & SSIM$\uparrow$ & LPIPS$\downarrow$ \\
\midrule
\multirow{4}{*}{Rectifier}
 & \multirow{2}{*}{Puzzle-IOI (10)}
 & \texttt{up2you} (host)        & 2.8400\gain{(-4.2\%)} & 1.7250\gain{(-8.6\%)} & 31.6146\gain{(-1.9\%)} & 23.4287\gain{(+4.3\%)} & 0.8487\gain{(+2.9\%)} & 0.1650\gain{(-12.4\%)} \\
 & & \textbf{Astrolabe@\texttt{up2you}}   & \textbf{2.7207} & \textbf{1.5770} & \textbf{31.0132} & \textbf{24.4269} & \textbf{0.8735} & \textbf{0.1445} \\
\cmidrule(lr){2-9}
 & \multirow{2}{*}{4D-Dress (10)}
 & \texttt{up2you} (host)        & \textbf{1.0760}\loss{(+0.5\%)} & 0.9872\gain{(-1.2\%)} & 25.4743\gain{(-0.2\%)} & 22.9239\gain{(+3.9\%)} & 0.8927\gain{(+1.0\%)} & 0.0761\gain{(-10.2\%)} \\
 & & \textbf{Astrolabe@\texttt{up2you}}   & 1.0810 & \textbf{0.9751} & \textbf{25.4270} & \textbf{23.8100} & \textbf{0.9017} & \textbf{0.0683} \\
\midrule
\multirow{6}{*}{Optimization}
 & \multirow{3}{*}{Puzzle-IOI}
 & AvatarBooth (n=8; non-paired) & 3.8467 & 3.9599 & 0.1648 & 21.9131 & 0.8615 & 0.2137 \\
 & & PuzzleAvatar (n=9)           & 1.4116\gain{(-8.5\%)} & 1.4175\gain{(-8.7\%)} & 0.0719\gain{(-22.4\%)} & 25.1297\gain{(+3.6\%)} & 0.9045\gain{(+0.6\%)} & 0.1288\gain{(-7.1\%)} \\
 & & \textbf{Astrolabe@PA} (n=9)  & \textbf{1.2917} & \textbf{1.2945} & \textbf{0.0558} & \textbf{26.0428} & \textbf{0.9103} & \textbf{0.1196} \\
\cmidrule(lr){2-9}
 & \multirow{3}{*}{4D-Dress (10)}
 & AvatarBooth                      & 4.5267 & 3.7508 & 41.9970 & 13.7951 & 0.7772 & 0.2970 \\
 & & PuzzleAvatar                    & \textbf{1.5221}\loss{(+1.8\%)} & 1.4890\gain{(-3.2\%)} & 31.2493\gain{(-9.3\%)} & 26.1198\gain{(+3.6\%)} & 0.8811\gain{(+2.6\%)} & 0.1198\gain{(-13.5\%)} \\
 & & \textbf{Astrolabe@PA}           & 1.5501 & \textbf{1.4408} & \textbf{28.3359} & \textbf{27.0671} & \textbf{0.9043} & \textbf{0.1036} \\
\bottomrule
\end{tabular}%
}
\end{table*}

\subsection{Implementation, datasets, and metrics}
\textbf{Hosts.} We bind the same frozen SPH correspondence cue to two diffusion systems. In \texttt{up2you}~\cite{up2you}, SPH supplies coarse scores to the existing reference router, while the spatial shift is enabled during attention-LoRA adaptation and joint denoising; the remaining backbone is frozen. PuzzleAvatar~\cite{xiu2024puzzleavatar} with SD~2.1 has no cross-view router, so it receives SPH through shift-matched personalization and shifted condition/readout during the native $10$k-step geometry and appearance optimization. Unless stated otherwise, rectifier inference uses $30$ denoising steps, initialization/readout scales of $0.7/0.5$, readout at $t>600$, and association blend $\beta=0.2$.

\textbf{Hyperparameters.} Both hosts use the fixed SPH$\to$shift map $M$, while scales and schedules follow their native regimes. The rectifier's effective shift is approximately $0.04$. Optimization personalization uses AdamW for $1$k embedding/decoder steps at $5{\times}10^{-4}$ and $4$k model-personalization steps at $2{\times}10^{-6}$ with weight decay $10^{-2}$; input-shift jitter follows the optimization binding in Sec.~\ref{sec:hosts}. The DMTet geometry/appearance phases use learning rates $10^{-3}/6.6{\times}10^{-4}$, classifier-free guidance $6.5/7.0$, and normalized timestep ranges $[0.04,0.30]/[0.02,0.98]$. In integer DDPM indices, geometry uses condition/readout supports $[40,300)/[80,300)$ and scale $0.2$ (maximum shift $0.16$); appearance uses $[0,1000)/[0,200)$, scale $0.2$, and multiplier $0.35$ (maximum shift $0.056$). One empirical association setting is reused across all rectifier data; each host-regime shift schedule is fixed across subjects and datasets, without per-subject tuning.

\textbf{Hardware.} Both bindings fit on a single $24$\,GB GPU without multi-GPU training. Their operating regimes differ: the rectifier performs one feed-forward multi-view pass, whereas the optimization host is trained separately for each subject, so absolute resource values are compared only within each host.

\textbf{Datasets.} We use \textbf{Puzzle-IOI}~\cite{xiu2024puzzleavatar}, an in-the-wild personal-album benchmark with substantial viewpoint, lighting, and occlusion variation, and \textbf{4D-Dress}~\cite{wang20244d}, which provides captured humans in diverse garments. The rectifier evaluation contains a fixed $10$-subject Puzzle-IOI subset with $12$ references per subject and $10$ exact-match 4D-Dress cases with benchmark GT scans; the optimization evaluation uses the paired nine-subject Puzzle-IOI set and $10$ 4D-Dress cases. Each rectifier 4D-Dress case provides eight partial observations: front, back, left, and right views split into upper- and lower-body crops. Puzzle-IOI target shape/pose is taken from GT17 fitting and rendered at the host's six slots ($0^\circ,45^\circ,90^\circ,180^\circ,270^\circ,315^\circ$). Each host follows its dataset-specific reconstruction and novel-view protocol. The two 4D-Dress examples in Figs.~\ref{fig:qual_main} and~\ref{fig:front_back_qual} are separate matched-input case studies, not part of the rectifier's ten-case aggregate.

\textbf{Metrics.} Within each comparison, prediction and GT RGBA are independently composited onto black and evaluated as full images using PSNR$\uparrow$, SSIM$\uparrow$, and scalar Alex-LPIPS$\downarrow$; foreground-spatial LPIPS is diagnostic only. For rectifier geometry, Chamfer is the bidirectional nearest-surface distance and P2S is the reconstruction-to-GT direction, both in cm. Normal$^{\dagger}$ is the unoriented angular surface-normal error in degrees and is used by the rectifier and the 4D-Dress optimization benchmark. Puzzle-IOI optimization instead uses the native unitless four-view normal-image error Normal$^{\ddagger}$; the two Normal scales are never compared numerically. The main comparison and rectifier ablation use separately frozen 2D export/evaluation runs, so repeated image rows are interpreted only through paired contrasts within their own table.

\textbf{Baselines.} Because Astrolabe changes correspondence guidance rather than the avatar representation, its primary controls are the same hosts without the adapter: \texttt{up2you} for rectification and PuzzleAvatar~\cite{xiu2024puzzleavatar} for optimization. These paired comparisons retain the reconstruction machinery and isolate the cue-to-host conversion; Sec.~\ref{sec:exp_injection} then factorizes its paths. AvatarBooth~\cite{zeng2023avatarbooth} is an available SDS reference, but its eight-subject Puzzle-IOI mean is non-paired and is not used for percentage changes or ranking. Direct numbers from TeCH~\cite{huang2024tech}, HaveFUN~\cite{Yang2024HaveFUN}, and PFAvatar~\cite{Xi2025PFAvatar} are omitted because their protocols are respectively single-image, not strictly unconstrained, or unavailable as a public implementation for matched evaluation. Effectiveness and portability claims therefore come from paired host contrasts.

\subsection{Weak-correspondence gains across hosts and datasets (C1, C3)}
\label{sec:exp_main}
Table~\ref{tab:t1_main} first shows that weak global correspondence improves rectified target views across both datasets. On Puzzle-IOI, Astrolabe@\texttt{up2you} improves all six metrics: Chamfer/P2S decrease by $4.2\%/8.6\%$, angular normal error by $1.9\%$, and PSNR/SSIM/LPIPS improve by $4.3\%/2.9\%/12.4\%$. On 4D-Dress, all image metrics improve, including a $10.2\%$ LPIPS reduction; P2S and normal error also decrease by $1.2\%$ and $0.2\%$, while Chamfer changes by only $0.5\%$. Thus the rectifier consistently benefits in view synthesis, with all Puzzle-IOI geometry metrics improving and 4D-Dress geometry remaining numerically stable overall.

The optimization results show that the same correspondence cue remains effective without an explicit reference router. Against paired PuzzleAvatar on Puzzle-IOI, Astrolabe@PA improves every metric, with the largest relative change in native normal error ($-22.4\%$). On 4D-Dress, it improves P2S by $3.2\%$, angular normal error by $9.3\%$, and all image metrics, while Chamfer changes by only $1.8\%$. AvatarBooth remains a non-paired reference and is not the source of these percentages. Together, the two host blocks support the portability of frozen SPH and its fixed conversion from weak global correspondence to host-native diffusion guidance.

\begin{table*}[!t]
\centering
\caption{\textbf{Front vs.\ Back view synthesis.} Black-background RGBA-composited full-image 2D metrics on Puzzle-IOI, separated into the front target slot ($0^\circ$) and back target slot ($180^\circ$). Percentages report the relative change achieved by Astrolabe only for paired or within-protocol baselines. AvatarBooth is an $n=8$ non-paired mean and therefore has no percentage annotation.}
\label{tab:front_back}
\setlength{\tabcolsep}{6pt}
\renewcommand{\arraystretch}{1.08}
\resizebox{\textwidth}{!}{%
\begin{tabular}{l ccc ccc}
\toprule
 & \multicolumn{3}{c}{\textbf{Front ($0^\circ$)}} & \multicolumn{3}{c}{\textbf{Back ($180^\circ$)}} \\
\cmidrule(lr){2-4}\cmidrule(lr){5-7}
\textbf{Method} & PSNR$\uparrow$ & SSIM$\uparrow$ & LPIPS$\downarrow$ & PSNR$\uparrow$ & SSIM$\uparrow$ & LPIPS$\downarrow$ \\
\midrule
\multicolumn{7}{l}{\textit{Optimization (SDS) Host}} \\
AvatarBooth (n=8; non-paired) & 20.6720 & 0.8245 & 0.2442 & 21.3448 & 0.8279 & 0.2480 \\
PuzzleAvatar & 24.5894\gain{(+2.0\%)} & 0.8855\gain{(+0.6\%)} & 0.1240\gain{(-7.5\%)} & 24.3683\gain{(+6.3\%)} & 0.8851\gain{(+1.1\%)} & 0.1328\gain{(-9.3\%)} \\
\textbf{Astrolabe@PA} & \textbf{25.0859} & \textbf{0.8904} & \textbf{0.1147} & \textbf{25.9027} & \textbf{0.8950} & \textbf{0.1204} \\
\midrule
\multicolumn{7}{l}{\textit{Rectifier Host}} \\
\texttt{up2you} & 23.3907\gain{(+6.5\%)} & 0.8396\gain{(+4.2\%)} & 0.1566\gain{(-18.8\%)} & 24.2639\gain{(+4.2\%)} & 0.8427\gain{(+4.2\%)} & 0.1698\gain{(-17.4\%)} \\
\textbf{Astrolabe@\texttt{up2you}} & \textbf{24.9059} & \textbf{0.8745} & \textbf{0.1271} & \textbf{25.2802} & \textbf{0.8782} & \textbf{0.1402} \\
\bottomrule
\end{tabular}
}
\end{table*}

\subsection{Difficult-view and qualitative analysis}
Rear views provide a direct stress test of global correspondence because their evidence is often sparse or observed elsewhere in the album. Table~\ref{tab:front_back} separates the $0^\circ$ and $180^\circ$ Puzzle-IOI targets. Relative to PuzzleAvatar, optimization improves back-view PSNR/SSIM by $6.3\%/1.1\%$ and reduces LPIPS by $9.3\%$, compared with $2.0\%/0.6\%/7.5\%$ on the front. Rectification is similarly consistent, reducing LPIPS by $18.8\%$ on the front and $17.4\%$ on the back. The gains therefore extend to the viewpoint where cross-view organization is most needed.

\begin{figure}[!b]
    \centering
    \includegraphics[width=\linewidth,clip,trim=3.15cm 9.0cm 1.35cm 2.4cm]{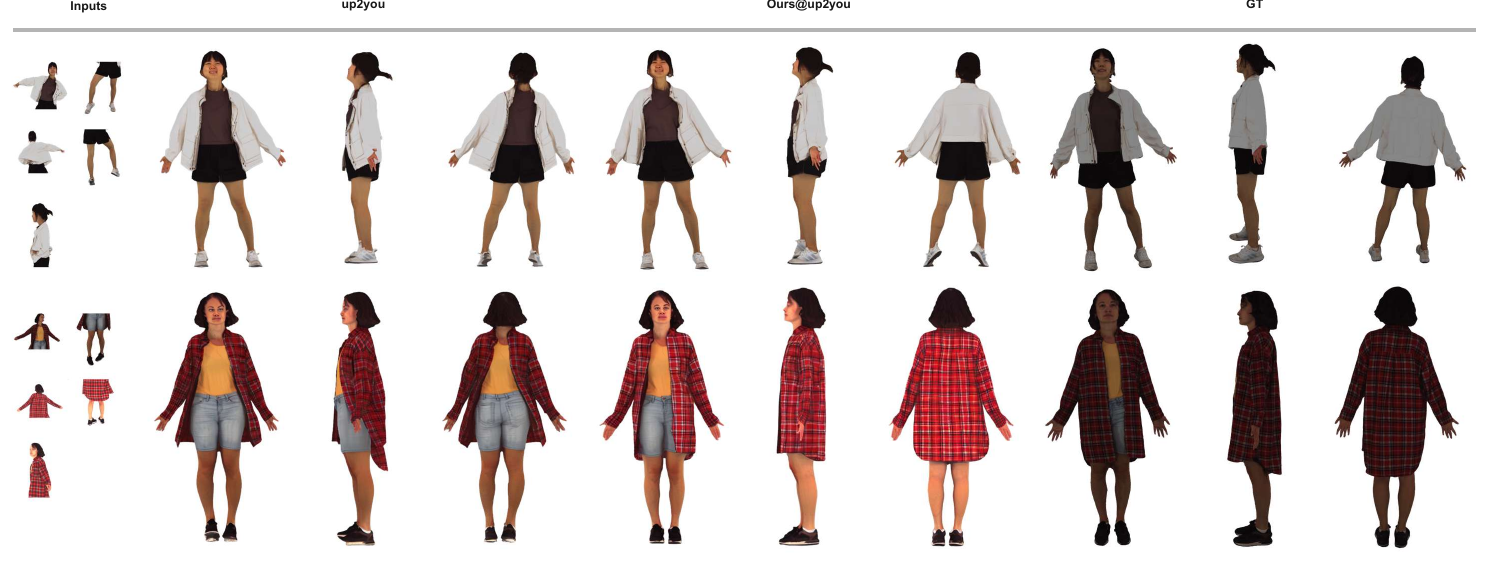}
    \caption{\textbf{Front/side/back rectifier comparison on two 4D-Dress cases.} Partial inputs are followed by \texttt{up2you}, Astrolabe@\texttt{up2you}, and ground truth at matched views, exposing rear-garment layout and silhouette recovery.}
    \label{fig:front_back_qual}
\end{figure}

Qualitative comparisons localize these gains to regions where correspondence is ambiguous. Relative to each host, Fig.~\ref{fig:qual_main} shows cleaner head and limb boundaries, more stable hair--neck transitions, and more coherent rear-garment extent and patterns. Matched RGB and normal views separate appearance continuity from surface behavior; the partial-observation cases in Fig.~\ref{fig:front_back_qual} further expose rear-jacket silhouette and plaid continuity. These 4D-Dress examples follow the separate matched-input case-study protocol in Fig.~\ref{fig:qual_main}; they illustrate rather than enlarge the ten-case aggregate.

\subsection{Factorizing weak-correspondence guidance (C2, C3)}
\label{sec:exp_injection}
Table~\ref{tab:t2_injection} separates the two ways that SPH correspondence enters the rectifier: coarse reference association and the coupled shift path. The two SPH-only controls contain association alone---no spatial shift during adaptation or inference and no shift-matched LoRA---and differ only in whether association values are raw or flattened. Rows that activate the shift path include its matched LoRA by construction; the table therefore exposes only its initialization and high-noise readout uses rather than presenting LoRA as an independent component. All variants use the Puzzle-IOI setup of $10$ subjects, six targets, $12$ references, matched seeds and denoising steps, and one frozen $768^2$ RGBA exporter/evaluator. Table~\ref{tab:t1_main} and this ablation use their respective frozen 2D export/evaluation runs, so conclusions are drawn from paired rows within Table~\ref{tab:t2_injection}.

\begin{figure}[!b]
    \centering
    \includegraphics[width=0.99\linewidth,clip,trim=0cm 12.7cm 0cm 5.3cm]{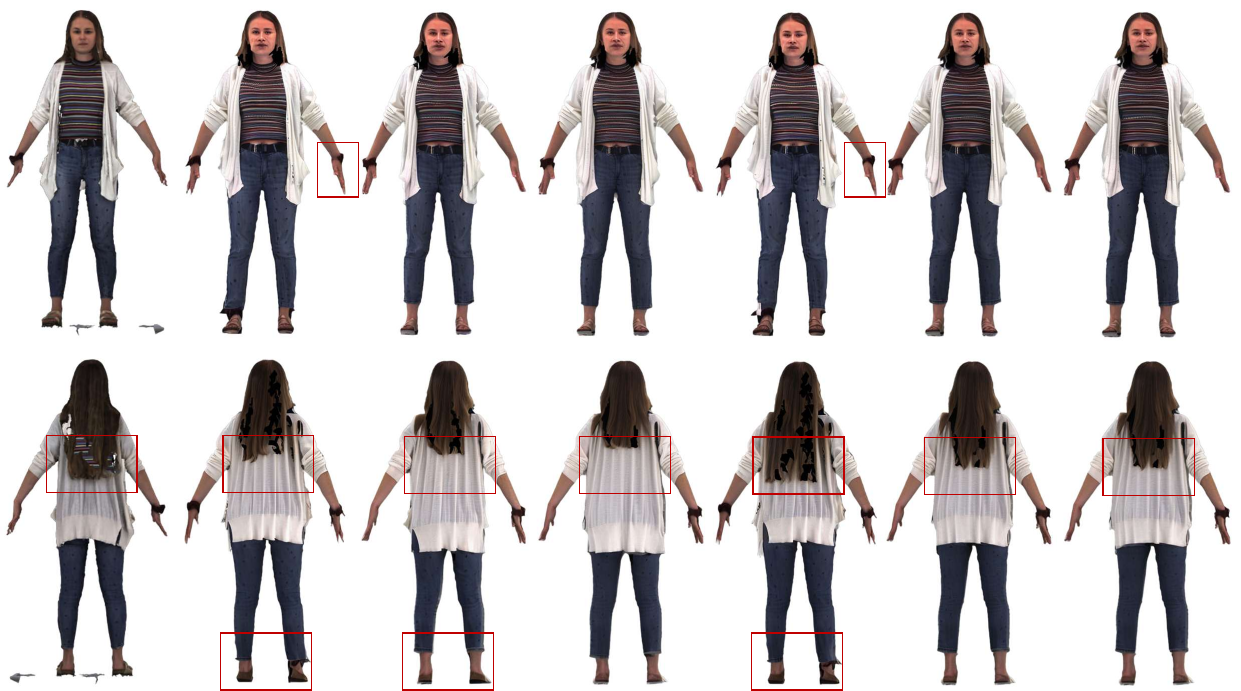}
    \caption{\textbf{Qualitative rectifier injection-path ablation corresponding to Table~\ref{tab:t2_injection}.} Front/back comparisons show the host, raw and flattened SPH-only association (both without shift or LoRA), the shift path without association, initialization-only guidance, and full Astrolabe. The highlighted regions emphasize hand/arm artifacts and back-garment consistency.}
    \label{fig:qual_rect_ablation}
\end{figure}

Association-only SPH supplies most of the measured rectifier improvement. Without either shift or LoRA, its raw and flattened variants both improve all three image metrics over the host and remove the highlighted hand/arm and rear-garment artifacts in Fig.~\ref{fig:qual_rect_ablation}. Adding the complete coupled shift path to flattened association further changes PSNR/SSIM/LPIPS from $24.3153/0.8841/0.1403$ to $24.4651/0.8854/0.1385$, a smaller but consistent improvement in all three metrics. Most of this complement is already present in the initialization-only row; adding high-noise readout changes PSNR by $0.0053$\,dB and leaves SSIM/LPIPS equal at four decimals. Conversely, the shift path without association does not reproduce the association gain. The rectifier evidence therefore supports coordinated roles: coarse compatibility drives native reference selection, while the coupled spatial shift provides the shared trajectory interface used across both pipeline categories.

\begin{table*}[!tbp]
\centering
\caption{\textbf{Factorized injection-path ablation (rectifier, Puzzle-IOI, 10 subj.).} SPH-only is association-only: it uses neither spatial shift nor LoRA. ``Raw'' and ``flat'' denote association with $\beta=1$ and $0.2$. Any active shift path includes its shift-matched LoRA by construction; Init./Readout indicate its two inference uses rather than separate adapters. All rows use one frozen $768^2$ RGBA evaluator, and bold marks the complete adapter.}
\label{tab:t2_injection}
\setlength{\tabcolsep}{5.2pt}
\renewcommand{\arraystretch}{1.04}
\resizebox{\textwidth}{!}{%
\begin{tabular}{l cc cc ccc}
\toprule
 & \multicolumn{2}{c}{\textbf{SPH Association}} & \multicolumn{2}{c}{\textbf{Coupled Shift Path}} & \multicolumn{3}{c}{\textbf{2D (Multi-view)}} \\
\cmidrule(lr){2-3}\cmidrule(lr){4-5}\cmidrule(lr){6-8}
\textbf{Variant} & Raw & Flat & Init. & Readout & PSNR$\uparrow$ & SSIM$\uparrow$ & LPIPS$\downarrow$ \\
\midrule
\texttt{up2you} (host)                    & \abno  & \abno  & \abno  & \abno  & 23.4644 & 0.8612 & 0.1633 \\
SPH-only (w/o shift), raw                 & \abyes & \abno  & \abno  & \abno  & 24.2285 & 0.8844 & 0.1392 \\
SPH-only (w/o shift), flattened           & \abno  & \abyes & \abno  & \abno  & 24.3153 & 0.8841 & 0.1403 \\
shift path only (w/o association)          & \abno  & \abno  & \abyes & \abyes & 23.4237 & 0.8477 & 0.1797 \\
init-only (no readout)                    & \abno  & \abyes & \abyes & \abno  & 24.4598 & 0.8854 & 0.1385 \\
\textbf{Astrolabe (complete)}             & \abno  & \abyes & \abyes & \abyes & \textbf{24.4651} & \textbf{0.8854} & \textbf{0.1385} \\
\bottomrule
\end{tabular}
}
\end{table*}

Table~\ref{tab:t3_optim} asks how the SPH correspondence field remains useful when the optimization host exposes no reference router. Two controls disable shift matching during personalization or guided NFDS readout during distillation. The channel-mean control replaces spatial $\delta(h,w)$ with foreground mean $\Delta$, preserving each channel shift while removing global layout. Two further variants discard latent shifting and inject the same online SPH map or a DINOv2 PCA map through a frozen ControlNet-color branch during both stages.

\begin{figure}[!b]
    \centering
    \includegraphics[width=\linewidth,clip,trim=5.45cm 3.35cm 6.2cm 2.2cm]{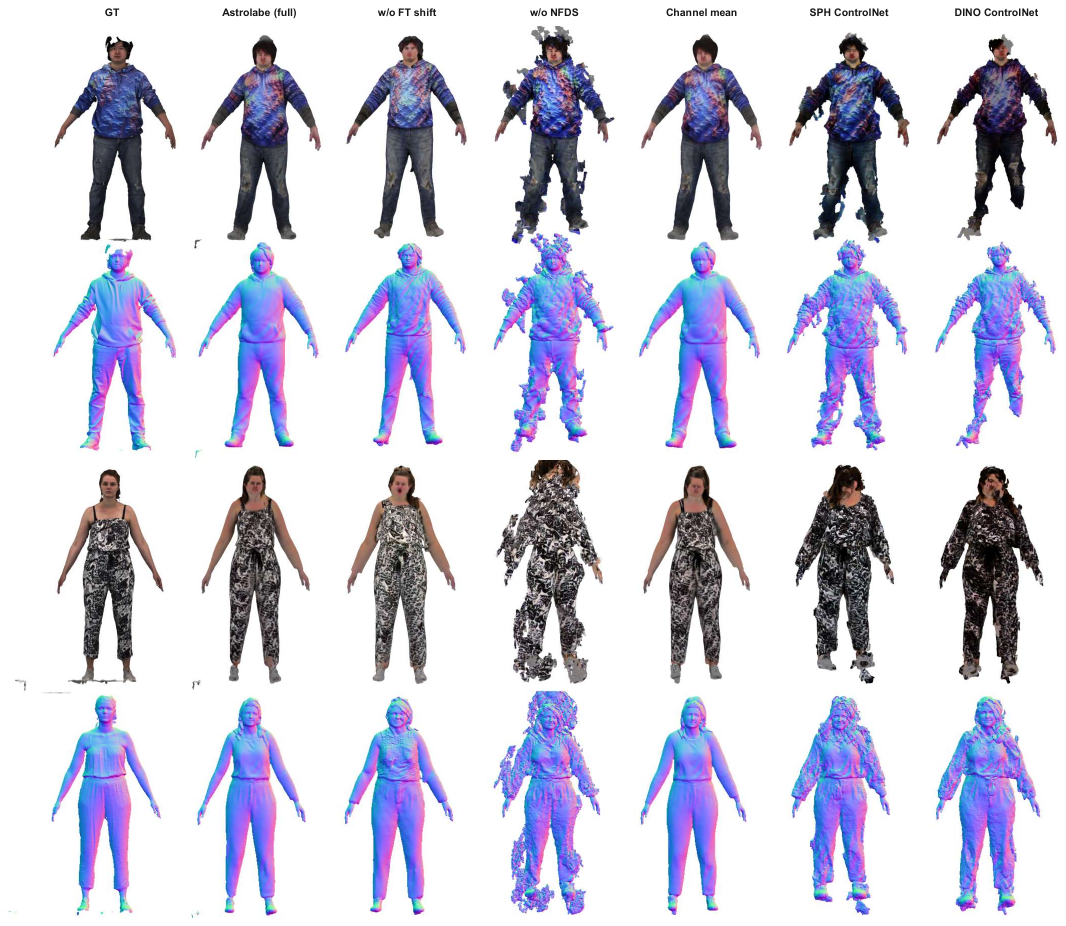}
    \caption{\textbf{Qualitative optimization-host mechanism ablation corresponding to Table~\ref{tab:t3_optim}.} RGB and normal renderings compare full Astrolabe, removal of shift-matching fine-tuning, removal of the guided NFDS readout, channel-mean guidance, and the SPH/DINO ControlNet alternatives.}
    \label{fig:qual_optim_ablation}
\end{figure}

Guided NFDS readout is the strongest measured path in the optimization binding. Removing it lowers PSNR/SSIM from $25.2784/0.9003$ to $20.1682/0.8094$, raises LPIPS from $0.1071$ to $0.2373$, and produces the fragmented reconstruction in Fig.~\ref{fig:qual_optim_ablation}. Removing shift matching gives a smaller but consistent image change: the complete adapter adds $1.5820$\,dB PSNR and $0.0171$ SSIM while reducing LPIPS by $0.0161$. Replacing spatial $\delta$ with channel mean $\Delta$ also reduces every image metric, directly showing that global layout, rather than a channel-wide bias alone, carries useful correspondence. Full Astrolabe further outperforms both frozen ControlNet-color alternatives on all three image metrics while avoiding their heavier branch (Table~\ref{tab:t5_efficiency}). Thus, without explicit association, weak global correspondence remains effective through matched spatial trajectory guidance and its NFDS readout.

\begin{table*}[!tbp]
\centering
\footnotesize
\caption{\textbf{Mechanism-transfer ablation (optimization host, Puzzle-IOI, 9 subj.).} Component columns identify the shift layout and the stages at which it is matched or read out. The last two rows replace the complete Astrolabe shift path with a frozen ControlNet-color branch that injects SPH or DINO during both personalization and distillation. All rows use black-background RGBA compositing and full-image scalar Alex-LPIPS.}
\label{tab:t3_optim}
\setlength{\tabcolsep}{4.7pt}
\renewcommand{\arraystretch}{1.04}
\resizebox{\textwidth}{!}{%
\begin{tabular}{l cc cc cc ccc}
\toprule
 & \multicolumn{2}{c}{\textbf{Shift Layout}} & \multicolumn{2}{c}{\textbf{Shift Coupling}} & \multicolumn{2}{c}{\textbf{ControlNet-color}} & \multicolumn{3}{c}{\textbf{2D (Rendered)}} \\
\cmidrule(lr){2-3}\cmidrule(lr){4-5}\cmidrule(lr){6-7}\cmidrule(lr){8-10}
\textbf{Variant} & Spatial $\delta$ & Mean $\Delta$ & FT Match & NFDS Readout & SPH & DINO & PSNR$\uparrow$ & SSIM$\uparrow$ & LPIPS$\downarrow$ \\
\midrule
\textbf{Astrolabe (complete)} & \abyes & \abno  & \abyes & \abyes & \abno  & \abno  & \textbf{25.2784} & \textbf{0.9003} & \textbf{0.1071} \\
w/o shift matching (FT)       & \abyes & \abno  & \abno  & \abyes & \abno  & \abno  & 23.6964 & 0.8832 & 0.1232 \\
w/o guided readout (NFDS)     & \abyes & \abno  & \abyes & \abno  & \abno  & \abno  & 20.1682 & 0.8094 & 0.2373 \\
channel $\Delta$              & \abno  & \abyes & \abyes & \abyes & \abno  & \abno  & 24.0154 & 0.8891 & 0.1189 \\
\midrule
ControlNet-color (SPH)        & \abno  & \abno  & \abno  & \abno  & \abyes & \abno  & 22.5886 & 0.8575 & 0.1692 \\
ControlNet-color (DINO)       & \abno  & \abno  & \abno  & \abno  & \abno  & \abyes & 21.8782 & 0.8568 & 0.1740 \\
\bottomrule
\end{tabular}%
}
\end{table*}

\subsection{Weak correspondence: selection, not loudness (C2)}
\label{sec:exp_selection}
\begin{figure}[!b]
    \centering
    \includegraphics[width=0.80\linewidth, clip, trim=8cm 3.5cm 7cm 2.2cm]{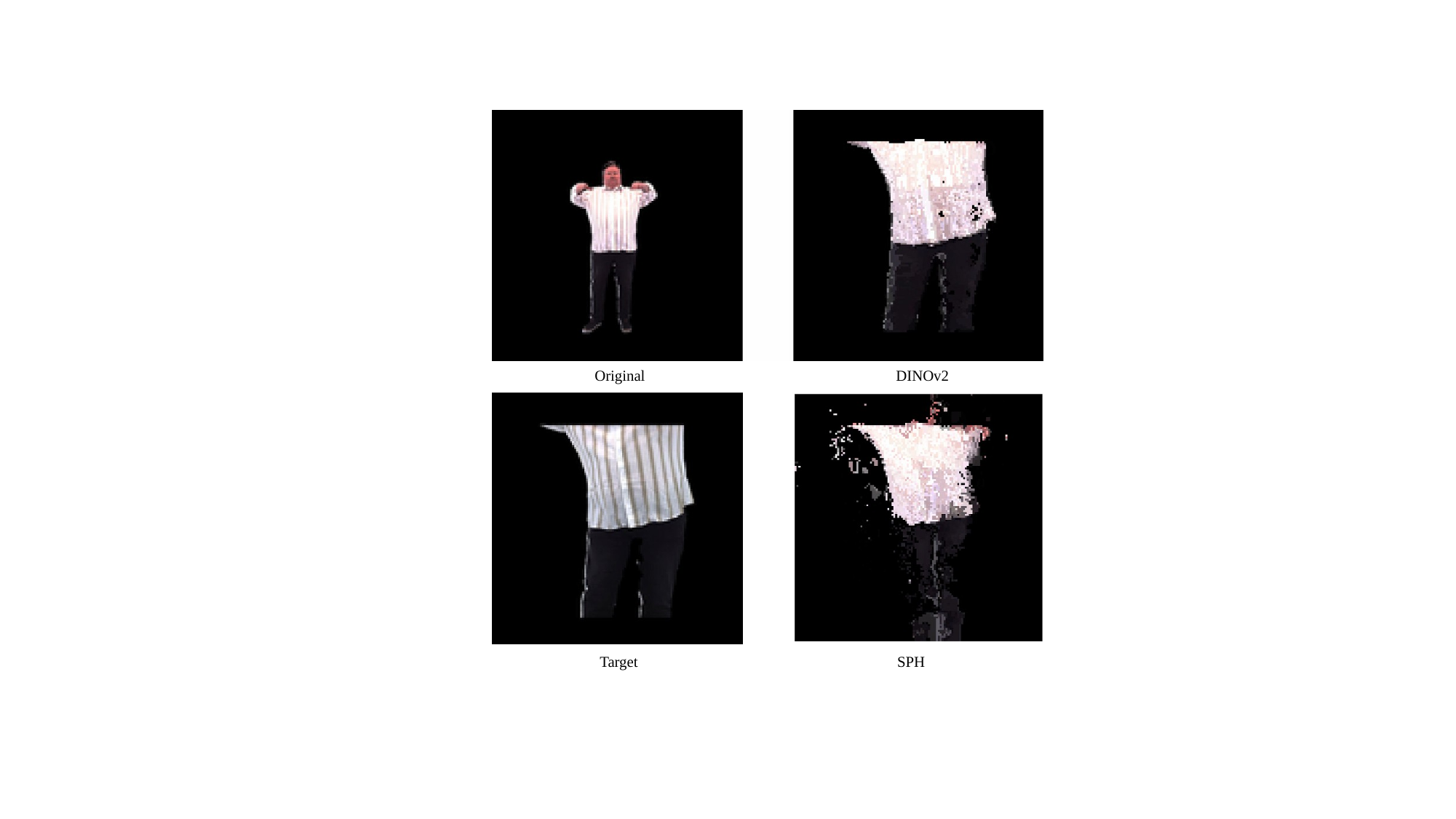}
    \caption{\textbf{Dense-correspondence stress test.} The left column shows the reference and target; the right column warps the reference with DINOv2 or SPH. Both transfers are distorted across the large viewpoint gap, yet coarse SPH association improves the host. Weak global compatibility, rather than accurate dense transfer, is sufficient.}
    \label{fig:correspondence}
\end{figure}

SPH should be interpreted as weak global correspondence rather than a dense matching field. Figure~\ref{fig:correspondence} warps a reference across a large viewpoint change using DINOv2 or SPH; both transfers are distorted and noisy. Yet SPH improves the rectifier when it only ranks and selects native reference features. Its useful content is therefore coarse cross-view compatibility, not accurate pixel-to-pixel transfer.

\begin{figure}[!b]
    \centering
    \includegraphics[width=0.97\linewidth, clip, trim=0cm 2cm 0cm 0cm]{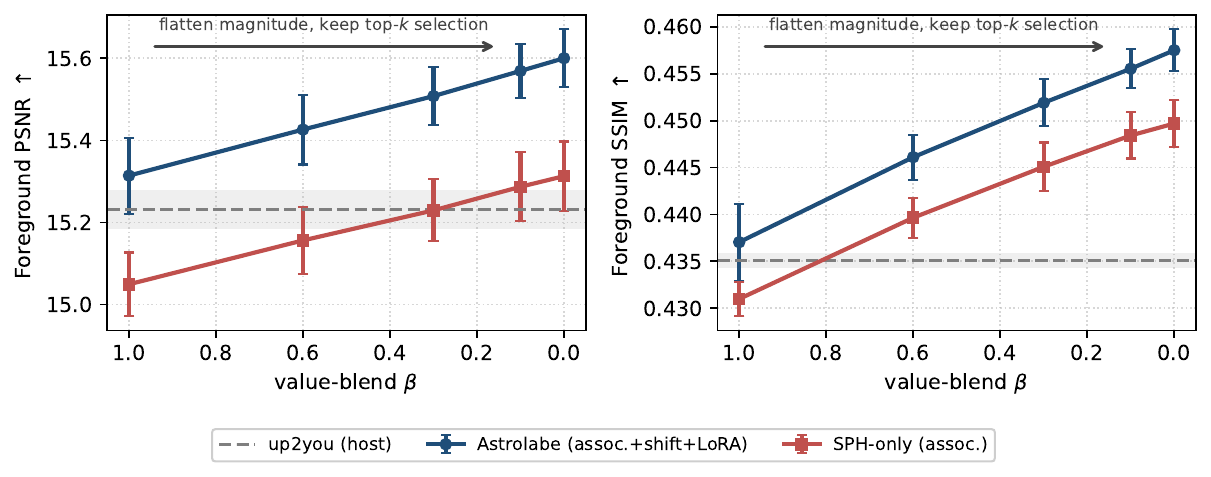}
    \caption{\textbf{Selection, not loudness} (rectifier, gt17-as-GT, 10 subjects). For $0<\beta\leq1$, flattening preserves support/ranking on the native SPH grid, although interpolation can alter boundary order before top-$k$; $\beta=0$ ties all positive values. Foreground-only PSNR/SSIM improve under compression. Curves and error bars show the mean and standard deviation over three seeds; the gray band shows the host standard deviation.}
    \label{fig:beta_alpha}
\end{figure}

Large association magnitudes are also unnecessary. In the foreground-only three-seed sweep of Fig.~\ref{fig:beta_alpha}, decreasing positive $\beta$ retains native-grid support/ranking while improving PSNR and SSIM for SPH-only association and the complete adapter. The operation compresses contrast rather than fixing the final selector exactly: feature-scale interpolation can alter boundary order, and $\beta=0$ ties all positive native values. Under the separate full-image protocol of Table~\ref{tab:t2_injection}, flattening SPH-only raises PSNR while SSIM is effectively unchanged ($0.8844$ vs.\ $0.8841$). The evidence therefore shows robustness to association compression, not monotonic improvement across protocols.

Together, the dense-transfer stress test, factorized controls, and compression sweep favor coarse support and ranking over precise warping or large scores. The evidence neither requires identical selected features after interpolation nor implies that every stronger compression improves every metric. This qualified robustness to coarse selection is what we call ``selection, not loudness.''

\subsection{Efficiency (C3)}
\label{sec:exp_efficiency}
Host-native correspondence guidance keeps rectifier inference essentially cost-neutral while reducing resident model memory. Table~\ref{tab:t5_efficiency} shows that Astrolabe replaces DINOv2-L appearance routing with a DINOv2-B-backed SPH predictor, fixed association, and a small LoRA; DINOv2-B predicts SPH but does not enter the association distance. Resident fp16 weights decrease by $14.7\%$, while runtime VRAM and step time change by only $+0.5\%$ and $+1.4\%$. These values are compared only within the rectifier regime, so ``cost-neutral'' does not mean cost-free.

In optimization, the host-native shift is lighter and more accurate than the tested ControlNet-color alternative receiving the same online SPH. It reduces model memory by $20.9\%$, runtime VRAM by $16.4\%$, and SDS-step time by $15.7\%$. Together with Table~\ref{tab:t3_optim}, this result shows that weak correspondence can be carried by the existing noise interface instead of a learned control branch.

\begin{table}[!tbp]
\centering
\caption{\textbf{Host-specific efficiency} (lower is better). Memory denotes resident fp16 weights. Rectifier timing is the $10$-subject, six-view, $12$-reference protocol with $30$ denoising steps, reported per step; optimization timing is one batch-$3$, $512^2$ SDS step including online SPH. Parentheses report the relative change achieved by Astrolabe within each host.}
\label{tab:t5_efficiency}
\setlength{\tabcolsep}{2.2pt}
\renewcommand{\arraystretch}{1.14}
\resizebox{\linewidth}{!}{%
\begin{tabular}{@{}l c c c@{}}
\toprule
\textbf{Method} & \textbf{Model mem.} & \textbf{Runtime VRAM} & \textbf{Step time} \\
\midrule
\multicolumn{4}{@{}l}{\textit{Rectifier}} \\
\texttt{up2you} &
$3.6506$ GB \gain{(-14.7\%)} &
$19.6213$ GB \loss{(+0.5\%)} &
$314.75$ ms \loss{(+1.4\%)} \\
\textbf{Astrolabe@\texttt{up2you}} &
$3.1134$ GB &
$19.7238$ GB &
$319.22$ ms \\
\midrule
\multicolumn{4}{@{}l}{\textit{Optimization}} \\
PA+ControlNet (online SPH) &
$3.4857$ GB \gain{(-20.9\%)} &
$4.1769$ GB \gain{(-16.4\%)} &
$57.99$ ms \gain{(-15.7\%)} \\
\textbf{Astrolabe@PA} &
$2.7572$ GB &
$3.4917$ GB &
$48.90$ ms \\
\bottomrule
\end{tabular}%
}
\end{table}

\subsection{Evidence summary}
\label{sec:discussion}
Taken together, the experiments support Astrolabe's central claim: global correspondence can guide unconstrained reconstruction even when it is too weak for dense transfer. In the rectifier, association-only SPH improves native reference selection and supplies most of the image gain, while the coupled shift path adds a smaller consistent complement. In optimization, where no router is available, the same cue acts through spatial trajectory layout, matched personalization, and guided NFDS readout. Cross-host improvements, rear-view results, and correspondence diagnostics therefore support one complete adapter that translates weak global correspondence according to the variables exposed by each diffusion pipeline.

\section{Limitations and Ethics}
\label{sec:limitations}
Astrolabe inherits the weaknesses of its cue. \emph{Dependence on SPH/segmentation:} thin structures and heavy occlusion receive weak guidance, leading to over-smoothed extremities; confidence-aware fusion of SPH with normals/depth could help. \emph{Magnitude is regime-specific:} although the mechanism is shared, the effective shift magnitude must be set per host (small and trained-in for the rectifier, larger for single-step SDS); we set it by regime rather than learn it. \emph{Appearance/lighting:} the optimization host's simple albedo model entangles material and shading, limiting relighting. \emph{Ethics:} avatar capture from casual photos raises consent and privacy concerns; we recommend opt-in usage, anonymization, and provenance tracking.

\section{Conclusion}
Astrolabe brings global correspondence into unconstrained full-body capture through one adapter. Frozen SPH becomes a bounded spatial shift shared by rectification and optimization; an existing rectifier router additionally uses coarse compatibility. Across Puzzle-IOI and 4D-Dress, Astrolabe improves all reported image metrics in both hosts and all paired Puzzle-IOI geometry metrics; image gains extend to rear views, while 4D-Dress geometry remains stable overall. Factorized controls show that weak correspondence can guide both router-based association and router-free optimization. Dependence on cue quality motivates confidence-aware fusion and temporal consistency.

\FloatBarrier
\clearpage
\bibliographystyle{IEEEtran}
\bibliography{tvcg}

\end{document}